\documentclass[journal]{IEEEtran}
\usepackage[usenames]{color}
\usepackage{pdfpages}
\usepackage{epsfig}
\usepackage{graphics}
\usepackage{float}
\usepackage{hyperref}
\usepackage{eqparbox}
\newcommand*\rot{\rotatebox{90}}
\usepackage[cmex10]{amsmath}
\usepackage{amsmath}
\usepackage{amssymb}
\usepackage{xcolor}
\usepackage{multirow}
\usepackage{cite}
\usepackage{array}
\usepackage{makecell}
\usepackage{pslatex} 
\usepackage{url}
\usepackage{balance}
\usepackage{caption}
\usepackage{lineno}
\usepackage{soul}
\usepackage{xcolor}
\usepackage{graphicx}  
\usepackage{setspace}
\usepackage{tikz}
\usepackage{hhline}
\usepackage{mathtools}
\usepackage[letterpaper]{geometry}
 \newcommand{\flogo}{\includegraphics[height=18pt]{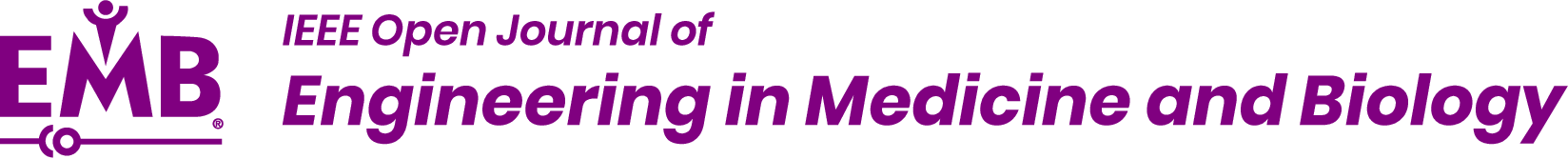}
 }
\ifCLASSINFOpdf
  
\else
  
\fi

\usepackage[english]{babel}
\usepackage[utf8]{inputenc}
\usepackage{fancyhdr}
\begin{document}

\title{\textcolor{violet}{Detecting of a Patient's Condition From Clinical Narratives Using Natural Language Representation\vspace{0.25cm}}}

\author{Thanh-Dung Le,~\IEEEmembership{Member,~IEEE,}
        Rita Noumeir Ph.D.,~\IEEEmembership{Member,~IEEE,}\\
      Jérôme Rambaud M.D., Ph.D., 
      Guillaume Sans M.D., and Philippe Jouvet M.D., Ph.D.
      % <-this % stops a space
\thanks{This work was supported in part by the Natural Sciences and Engineering Research Council (NSERC), in part by the Institut de Valorisation des données de l’Université de Montréal (IVADO), in part by the Fonds de la recherche en sante du Quebec (FRQS), and in part by the Fonds de recherche du Québec – Nature et technologies (FRQNT). 
	\par Thanh-Dung Le is with the Biomedical Information Processing Lab, \'{E}cole de Technologie Sup\'{e}rieure, University of Qu\'{e}bec, Canada, and also with the Research Center at CHU Sainte-Justine Hospital, University of Montreal, Canada (Email: thanh-dung.le.1@ens.etsmtl.ca).
	\par  Rita Noumeir is with the Biomedical Information Processing Lab, \'{E}cole de Technologie Sup\'{e}rieure, University of Qu\'{e}bec, Canada. 
	\par Jérôme Rambaud, Guillaume Sans, and Philippe Jouvet are with the Research Center at CHU Sainte-Justine Hospital, University of Montreal, Canada.
}}

\maketitle\thispagestyle{fancy}

\begin{abstract}
The rapid progress in clinical data management systems and artificial intelligence approaches enable the era of personalized medicine. Intensive care units (ICUs) are ideal clinical research environments for such development because they collect many clinical data and are highly computerized. \textit{Goal:} We designed a retrospective clinical study on a prospective ICU database using clinical natural language to help in the early diagnosis of heart failure in critically ill children. \textit{Methods:} The methodology consisted of empirical experiments of a learning algorithm to learn the hidden interpretation and presentation of the French clinical note data. This study included 1386 patients' clinical notes with 5444 single lines of notes. There were 1941 positive cases (36\% of total) and 3503 negative cases classified by two independent physicians using a standardized approach. \textit{Results:} The multilayer perceptron neural network outperforms other discriminative and generative classifiers. Consequently, the proposed framework yields an overall classification performance with 89\% accuracy, 88\% recall, and 89\% precision. \textit{Conclusions}: This study successfully applied learning representation and machine learning algorithms to detect heart failure in a single French institution from clinical natural language. Further work is needed to use the same methodology in other languages and institutions.
\end{abstract}


\IEEEpeerreviewmaketitle

\textbf{\textit{Impact Statement-} The study is a showcase to confirm that, dealing with a small dataset of clinical notes, a multilayer perceptron neural network classifier is a better approach compared to conventional classifiers, especially, pretrained-based deep learning models. Additionally, instead of losing information from numeric values, they can be retained and encoded for the representation learning. Consequently, it achieves better results for the classification task.}

\section{INTRODUCTION}

\IEEEPARstart{C}{urrently}, clinical narratives are continuously provided and stored in electronic medical records (EMR), but they are underutilized in clinical decision support systems. The limitation comes from their unstructured or semi-structured format. Besides, another problem with clinical narratives is that they are written in incomplete sentences but in an information-dense way for communication between clinicians \cite{johnson2016machine}. Because of the two reasons, clinical narrative sources impose constraints in an actual application for clinical outcome prediction.

Since 2013, the Pediatric Critical Care Unit at CHU Sainte-Justine (CHUSJ) has used an EMR. The patients' information, including vital signs, laboratory results, and ventilator parameters are updated every 5 minutes to 1 hour \cite{matton2016databases}. Primarily, a significant data source of French clinical notes is currently stored. There are seven caregiver notes/patient/day from 1386 patients (containing a dataset of more than $2.5\times 10^7$ words). These notes are scribed extensively from admission notes and evaluation notes. Admission notes outline reasons for admission to intensive care units, historical progress of the disease, medication, surgery, and the patient's baseline status. Daily ailments and test results are described in evaluation notes, from which patient condition is evaluated and diagnosed later by doctors. However, these information sources are being used as documentation for reporting and billing instead of clinical knowledge for predicting conditions or decision support. 

\subsection{Problem Statement}

The diagnosis of acute respiratory distress syndrome (ARDS) is frequently delayed or even not diagnosed in intensive care units. In the largest international cohort of patients with ARDS, the diagnosis of ARDS was delayed or missed in two-thirds of patients, with the diagnosis missed entirely in 40\% of patients \cite{bellani2016epidemiology}. To make the diagnosis of ARDS, three main conditions need to be detected: hypoxemia (low blood oxygenation), presence of infiltrates on chest X Ray and absence of cardiac failure \cite{pediatric2015pediatric}. The development of a clinical decision support system (CDSS) in real time that automatically screen the EMR data, chest X Rays and other data sources (medical devices collecting vital signs, ventilator settings) has the potential to increase diagnosis rate and then improve the management of this syndrome \cite{pediatric2015pediatric}. Our research team has developed the first two algorithms for hypoxemia \cite{sauthier2021estimated} and chest X Ray analysis \cite{zaglam2014computer}. This work contributes to the third algorithm development i.e. identifying the absence of cardiac failure. 

Cardiac failure is clinically suspected and the test that confirms its absence or presence is ususally an echocardiography. This echocardiography could have been performed prior to PICU admission, even in another institution and could not be digitally available for analysis. However, when an echocardiography has been performed, physicians report its result in the notes. It is the reason why, using notes to exclude or confimed a cardiac failure was assumed to be the best way to electronically collect as soon as possible the information.

Generally, there is a list of golden indicators to classify cardiac failure patients.  Those indicators could be either from the medical history, clinical exam, chest X-Ray interpretation, recent cardiovascular performance evaluation, or laboratory test results. Medication, such as Levosimendan, Milrinone, Dobutamine, is a surrogate to the gold standard. Its list can be retrieved from syringe pump data, prescriptions, and notes. If any medication from the three is present, there is certainly a cardiac failure. Besides, cardiovascular performance evaluation also contributes to indicate the cardiac failure diagnosis. One of the evaluations is ejection fraction (EF) $<50\%$. EF refers to the percentage of blood pumped (or ejected) out of the ventricles with each contraction. It is a surrogate for left ventricular global systolic function, defined as the left ventricular stroke volume divided by the end-diastolic volume. The other indicator for cardiovascular performance evaluation is shortening fraction (SF) $<25\%$. FR is the length of the left ventricle during diastole and systole. It measures diastolic/systolic changes for inter-ventricular septal and posterior wall dimensions. Finally, brain natriuretic peptide, known as pro-BNP ng/L $>1000$, comes from laboratory test results being useful in the acute settings for differentiation of cardiac failure from pulmonary causes of respiratory distress. Pro-BNP is continually produced in small quantities in the heart and released in more substantial quantities when the heart needs to work harder. 

Consequently, the clinical knowledge representation will summarize detailed attributes that are essential to detecting cardiac failure. All notes are taken into account if they are encompassed by the information of the prescription history of Milrinone (mcg/kg/min), measurement notes of pro-BNP (ng/L), dilated cardiomyopathy, acute left cardiac failure, chronic cardiac failure, postoperative cardiac failure, coronary microvascular disorder history notes, notes of a measurement result of either EF (\%) or SF (\%). As a result, a patient is considered to have a cardiac failure if he/she has one of the criteria. Unfortunately, as all the mentioned information above that helps diagnose cardiac failure is not readily available electronically, we will develop a machine learning algorithm based on natural language processing (NLP) that automatically detects this desired concept label from clinical notes. The algorithm can automatically see whether a patient has a cardiac failure or a healthy condition lacking gold indicators from the notes. In such a situation, the proposed algorithm can effectively learn a latent representation of clinical notes, which traditionally rule-based approaches cannot depict. 

\subsection{Motivation}

The recent study \cite{olsen2020clinical} extensively analyzed and confirmed the feasibility of employing machine learning for cardiac failure. However, we are dealing with two challenges from clinical notes in French and a limited amount of dataset size in our case. We will examine data retrospectively to validate the diagnosis. And the main objective of this study consists of two sub-objectives that overcome the mentioned limitations, as follows:

\begin{itemize}
    \item  Which representation learning approach should be used? The representation learning approach, which can retain  the  words’  semantic  and  syntactic  analysis  in critical care data, enriches the mutual information for the word representation by capturing word-to-word correlation. 
    \item  Which machine learning classifier should be employed? The classifier can avoid the overfitting associated with the machine learning rule by marginalizing over the model parameters instead of making point estimates of its values.
\end{itemize}

\section{MATERIALS AND METHODS }
\subsection{Clinical Narrative Data at CHUSJ}
\label{sec:data_CHUSJ}

\begin{figure}[t]
			\centering
			\includegraphics[width=0.375\textwidth]{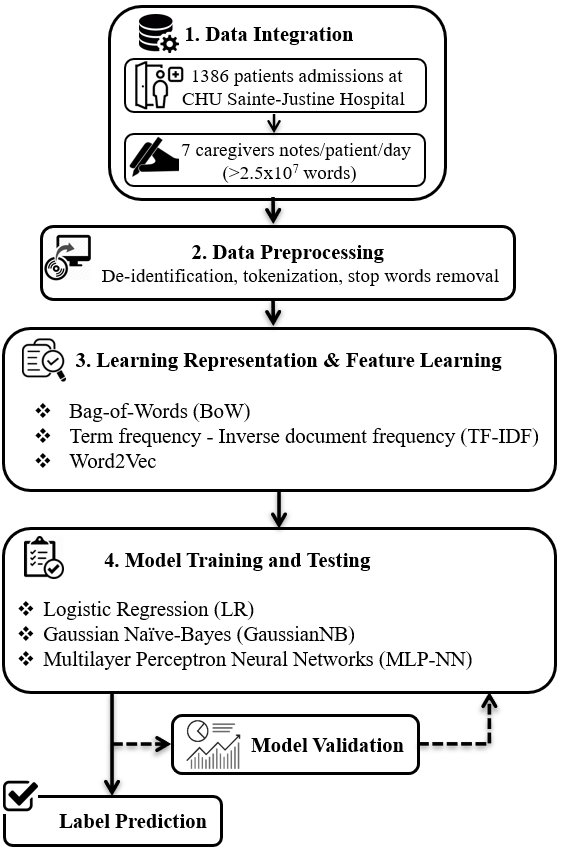}
			\caption{An overview of the proposed methodology to detect cardiac failure from clinical notes at CHUSJ.}
			\label{fig:method}
			\vspace{-5mm}
\end{figure}

Fig. \ref{fig:method} illustrates the conceptual framework for conducting the experiments. First, the data integration process has been completed at the Pediatric Intensive Care Unit, CHUSJ, for more than 1300 patients. After the research protocol was approved by the research ethics board from Research Center of the Sainte-Justine University Hospital. We only took information from two types of notes, including admission and evaluation notes. Since, these notes documented the reasons why a patient was admitted to the hospital by the physician in charge. And, the notes also provided the initial instruction for that patient's care based on the patient's health status. Primarily, we focused on medical background, history of the disease to admission, and cardiovascular evaluation. Furthermore, we only used notes for each patient's first stay within the first 24h since the admission. If a patient had more than one ICU stay, we only analyzed the first one. We did not have any missing notes but we can not exclude some information that were not collected by physicians and then not reported in the note. However, the data fully reflect real clinical practice. Then, two doctors from the CHUSJ (Dr. Jérôme Rambaud and Dr. Guillaume Sans), who did not compose the notes at the first hand, separately reviewed each patient's notes; each note was manually labeled ``YES" or ``NO" for positively cardiac failure or under a healthy condition, respectively. By doing so, we could double-check that missing data was not problematic. To avoid data contamination, we checked both the “patientID” and “careproviderID” to ensure no notes were simultaneously present in the training and testing cohort. Finally, we have 5444 line of notes with 1941 positive cases (36\% of total) and 3503 negative cases. The average length of the number of characters is 601 and 704. The average length of the number of digits is 25 and 26 for the positive and negative cases, respectively.

\subsection{Data Pre-Processing}
\label{sec:data_prep}
Generally, it is proven that if the preprocessing steps are well prepared, the result for the end-task will be improved \cite{kannan2014preprocessing}. Therefore, there are steps that were used as case lowering, and stop words removing. From the list, all these words are the definition in French; so, they do not contribute to the learning representation. Besides, we did not consider any French linguistic feature as our method is based on uncorrelated words. The longest n-gram is over 400 words, but most of the n-gram length distribution is between 50 and 125 words. %Then, the ratio of the number of samples/words per sample is much smaller than 1500, as given by a tutorial for small text classification \cite{Google_text_clf}.

In addition, it is essential to pay attention to negation in medical expression. First, the negation criteria from the study \cite{deleger2012detecting} were used for detecting the negative meaning from French notes. Then, a negation technique is applied  \cite{dubois2017learning}: a term ``neg\_" is added as a prefix for a term. An example note is ``Patient explique qu'à ce moment là, il \textit{n'était pas} capable de parler et l'air \textit{ne passait pas} au niveau de sa gorge. Respiration plus rapide, mais état général préservé, parents \textit{n'étaient pas} inquiets. (Patient explained at that time, he was not able to speak and the air did not pass at the level of his throat. Breathing faster, but general condition preserved, parents were not
worried)". The negation will be tagged as: ``Patient explique qu'à ce moment là, il \textbf{\textit{neg\_était}} capable de parler et l'air \textbf{\textit{neg\_passait}} au niveau de sa gorge. Respiration plus rapide, mais état général préservé, parents \textbf{\textit{neg\_étaient}} inquiets." 

\begin{figure*}[t]
			\centering
			\includegraphics[width=0.6\textwidth]{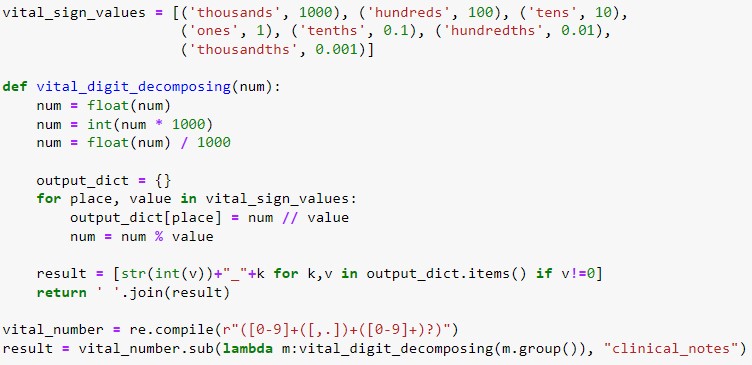}
			\caption{An example of code snippet in Python for decomposing numeric values (Example 4).}
			\label{fig:code_snippet}
			\vspace{-5mm}
\end{figure*}

For the  vital numeric values (heart rate, blood pressure, etc...), most of the NLP representation learnings cannot accommodate the numeric values effectively. Most NLP models treat numeric values in the text the same way as other tokens. It has been proven that the pre-trained token representations (word2vec) can naturally encode the numeric values  \cite{wallace2019nlp}. Unfortunately, it required a large amount of data with specific labeling progress for this task. At the same time, the state-of-the-art for numerical reasoning results is much less good (47\%) compared with the expert human performance (96.4\%) in the f1 score metric \cite{dua2019drop}. Another study only focuses on how to extract the number, not dealing with representation learning \cite{cai2019extraction}. Even, study \cite{kumar2020ensembling} proposes an alternative approach to deal with both large and small datasets. However, the authors either removed all of the vital sign numeric values or did not mention how to deal with numeric values. Because we have limited data, we decide to keep all numeric values for vital sign values (nearly 4\% of the notes) and apply the decoding for those number values. In fact, a numeric value consists in a numerical measurement value and a measurement unit as ruled by Digital Imaging and Communication in Medicine standard for report document \cite{noumeir2003dicom}. Therefore, we performed four experiments to evaluate the contribution from the numeric value to the classifiers. Fig. \ref{fig:code_snippet} shows an example of code snippet in Python, which help us conducting the decomposing the numerical measurement value. Finally, Table \ref{tab:numeric_decoding} summarizes the four different approaches to decode the numeric values, including (i) keeping all of the original numeric values and their units, (ii) removing all of the numeric values and their units, (iii) encoding the decimal into a string named dot, and (iv) decomposing into digits. 

\begin{table*}[t!]
\small
\centering
\caption{A summary of experiments dealing with vital sign numeric values.}
\label{tab:numeric_decoding}
\resizebox{\textwidth}{!}{\begin{tabular}{|c||c||c|}
\hline
Experiment  & Description & Illustration$^{*}$ \\    \hline

Exp\_1 &  Keep all of the numeric values and units & {[}vg, sévèrement, dilate, 64.8, mm, diastole, 58.3, mm, systole] \\    \hline

Exp\_2 & Remove all of the numeric values and units & {[}vg, sévèrement, dilate, diastole, systole] \\    \hline

Exp\_3 & Encoding the decimal point into string (DOT) & {[}vg, sévèrement, dilate, 64, dot, 8, mm, diastole, 58, dot, 3, mm, systole]\\    \hline

Exp\_4 & Decomposing numeric values into digits & \makecell{[vg, sévèrement, dilate, 6\_tens, 4\_ones, 8\_tenths, mm, \\ diastole, 5\_tens, 8\_ones, 3\_tenths, mm, systole]} \\   \hline

\multicolumn{3}{|l|}{$^{*}$The original notes are ``VG sévèrement dilaté (64.8mm en diastole et 58.3mm en systole) - Severely dilated LV (64.8mm in diastole and 58.3mm in systole)”} \\ \hline
\end{tabular}}
\vspace{-1mm}
\end{table*}

\subsection{Clinical Natural Language Representation Learning}
\label{sec:cli_nlp_learning}
There is no doubt about the effectiveness of neural word embedding. The study \cite{shi2016multiple} confirms that word2vec representation has been successfully used for various disease classifications from medical notes. Especially for the French clinical notes, the study \cite{dynomant2019word} shows that word2vec and GloVec effectively embed the clinical notes. And, the word2vec had the highest score on 3 out of 4 rated tasks (analogy-based operations, odd one similarity, and human validation). In addition, studies \cite{agarwal2017natural, li2018early, zhang2017automatic, fodeh2019classification} confirm that conventional approaches bag-of-words (BoW), term frequency-inverse document frequency (TF-IDF) have better performance than other deep learning techniques on a smaller corpus with long texts in clinical note corpus. Therefore, we will evaluate the effectiveness of two conventional representation approaches, including BoW, TF-IDF, and the word2vec neural embedding model.

\subsection{Machine Learning Classifiers}
\label{sec:ml_classifiers}

The state-of-the-art machine learning-based NLP currently focuses on deep learning for clinical notes \cite{pham2017predicting, rajkomar2018scalable, otter2020survey, young2018recent, sheikhalishahi2019natural}. For example, to predict cardiac failure, deep learning (Convolution Neural Network-based) shows its exceptional performance, F1 score of 0.756, to the conventional approach Random Forest (RF) with an F1 score of 0.674  \cite{liu2019predicting}. And, study \cite{shi2016multiple} shows the best performance to predict multiple chronic diseases (cerebral infraction, pulmonary infection and coronary atherosclerotic heart disease)  by combining of word2vec and deep learning with the average accuracy and F1 score exceeded 90\%. 

However, a large enough amount of data is needed to have a good generalization capability of deep learning, while this data availability requirement is not always provided \cite{paleyes2020challenges}. Especially, clinical notes in a language other than English, the challenge is more difficult to mitigate \cite{neveol2018clinical}. Deep learning architectures generally work well for large scale data sets with short texts while do not outperform conventional approaches (BoW) on a smaller corpus with long texts in clinical note corpus \cite{li2018early}. Automatic methods to extract New York heart association classification from clinical notes \cite{zhang2017automatic} confirm that the machine learning method, support vector machines (SVM) with n-gram features, achieves the best performance at 93\% F-measure. Also, study \cite{agarwal2017natural} proved the achievement by combining the BoW and Na\"ive Bayes classifier on clinical notes for accessing hospital readmission offering an area under the curve (AUC) of 0.690. This study confirms that, with the small dataset, TF-IDF and BoW have better performance than other techniques on coronary microvascular classification \cite{fodeh2019classification}. 

Besides, logistic regression (LR) and
generative Na\"ive Bayes perform better than the other
classifiers, particularly for small datasets. Several classifiers have been trained for short text classification; it includes RF, Gaussian Na\"ive Bayes (GaussianNB), Multinomial Na\"ive Bayes (MultimonalNB), LR, SVM and K-nearest neighbour. The experimental results from \cite{maimon2014data, wang2017comparisons} confirm that  LR, and GaussianNB perform much the better than the other classifiers. Moreover, study \cite{jordan2002discriminative} evaluated different classifiers' performance, including discriminative and generative learning approaches. And, it also confirms that the discriminative LR algorithm has a lower asymptotic error, while the generative Na\"ive Bayes classifier converges quickly.  

Additionally, when the ratio value for the number of samples/number of words per sample is small ($<$ 1500), a small multilayer perceptron neural network (MLP-NN) that takes n-grams as input performs better or at least as well as deep learning models \cite{Google_text_clf}. Besides, an MLP-NN is simple to define and understand, and it takes less computation time than sequence models. A detailed explanation of using an MLP-NN in medical analysis can be seen from \cite{pasini2015artificial}.

Consequently, we implemented and compared all the above mentioned methods; the result of RF, MultimonalNB, and SVM was less than 75\% for accuracy. Again, the result shows that only LR, GaussianNB and MLP-NN are comparable, and perform better than RF, MultimonalNB, and SVM classifer. Therefore, in this study, we focus on three different machine learning classifiers, including LR, GaussianNB, and MLP-NN.

\section{Results}
\label{sec:result}
We did the analysis to select of proper neural network sizes and architectures \cite{hunter2012selection}. We have used the structure of an MLP-NN that consists of $L=3$ layers, where layer 1 is the input layer, layer 3 is the output layer, and layer 2 is the hidden layer. The total number of neurons in the hidden layer is $N_t = 100$ neurons. To prevent the neural network from overfitting, we applied the dropout \cite{srivastava2014dropout} with the probability of dropping out rate p=0.25, and GlorotNormal kernel initializer \cite{glorot2010understanding}.

We used the scikit-learn library \cite{scikit-learn} and Keras \cite{chollet2015keras} in Python to implement our model. No preprocessing was required to deal with missing data. The data was divided into 60\% training, 20\% validation, and 20\% testing. To make our results more consistent, we used the $k$-fold cross validation ($k=5$) \cite{kohavi1995study}; each dataset was divided into $k$ subsets called folds, the model was trained on $k-1$ of them and tested on the left out. This process was repeated $k$ times, and the results were averaged to get the final one. Furthermore, we also employed the univariate feature selection with sparse data from the learning representation feature space. This selection process works by selecting the best features based on univariate statistical tests named SelectKBest algorithms, which removes all but the  $K$ highest scoring features (K=20000).

To effectively assess the performance of our method, metrics including accuracy, precision, recall (or sensitivity), and F1 score were used \cite{goutte2005probabilistic}. These metrics are defined as follows: 
\begin{align}
&\text {Accuracy (acc) }=\frac{\mathrm{TP}+\mathrm{TN}}{\mathrm{TP}+\mathrm{TN}+\mathrm{FP}+\mathrm{FN}} \nonumber \\ 
&\text {Precision (pre) }=\frac{\mathrm{TP}}{\mathrm{TP}+\mathrm{FP}} \nonumber \\
&\text {Recall/Sensitivity (rec)}=\frac{\mathrm{TP}}{\mathrm{TP}+\mathrm{FN}} \nonumber \\ 
&\text {F1-Score (f1)} =\frac{2^{\star} \text {Precision}^{\star} \text {Recall}}{\text {Precision }+\text {Recall}} \nonumber
\end{align}

\noindent where TN and TP stand for true negative and true positive, respectively, and they are the number of negative and positive patients classified correctly. FP and FN represent false positive and false negative, respectively, representing the number of positive and negative patients wrongly predicted. 

\section{Discussion}
\label{sec:discussions}

Table \ref{tab:table-results} presents the results of our method. First, among four experiments for dealing with numeric values, experiment 3 yields the best performance. Encoding the decimal point into a string ``DOT" has helped the learning representation process retain the information from numeric values. It is also interesting to mention that when we keep all numeric values and do nothing (experiment 1), the results are worse than if we remove all the numbers and their units (experiment 2). Experiment 4 confirms that if the numbers are extensively encoded, it will negatively affect the result, lowering the performance.

\begin{figure}[!tp]
	\centering
	\includegraphics[scale=0.6]{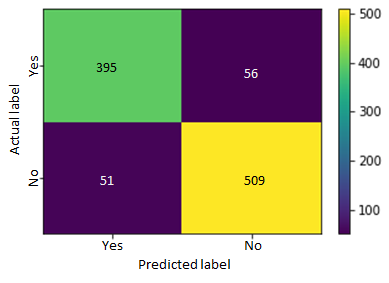}
	\caption{Confusion matrix of the MLP-NN classifier, showing the classification of positive (Yes) and negative (No) between predicted and actual labels.}
	\label{fig:confusion_matrix}
	\vspace{-2mm}
\end{figure}

The combination of TF-IDF and MLP-NN consistently outperforms other combinations with overall performance and is the most stable in all circumstances. Without any feature selection, the proposed framework yielded an overall classification performance with acc, pre, rec, and f1 of 85\% and 84\%, 85\%, and 84\%, respectively. Also, the representation matrix from the TF-IDF above is sparse because every word is treated separately. Hence, the semantic relationship between separated entities is ignored, which would cause information loss. Therefore, if the feature selection (SelectKBest) was well applied and tuned, it could improve up to 3-4\% for each evaluation in the overall performance. Consequently, it achieves the best performance with 89\%, 89\%, 88\%, and 88\% for acc, pre, rec, and f1, respectively. And, the detailed confusion matrix showing the classification of positive cases (1) and negative cases (0) is shown in Fig. \ref{fig:confusion_matrix}.

\begin{figure}[!tp]
	\centering
	\includegraphics[scale=0.67]{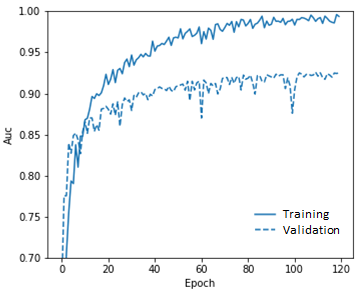}
	\caption{Area Under the Curve (AUC) performance of MLP-NN.}
	\label{fig:AUC}
	\vspace{-5mm}
\end{figure}

\begin{figure*}[!tp]
	\centering
	\includegraphics[scale=0.56]{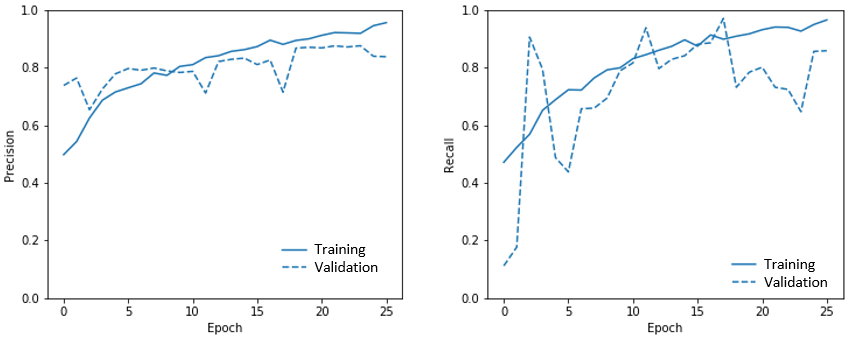}
	\caption{Precision (left) and recall (right) performance based on the Transformer configuration.}
	\label{fig:trans_recall}
	\vspace{-2mm}
\end{figure*}

Furthermore, with limited data, the BoW and TF-IDF have proven their capacity to better retain information from the notes representation. It has been shown in \cite{wang2017comparisons} that the TF-IDF has the highest accuracy compared to neural word embeddings in short text classification (less than 20 words per sample). In our study, we could not increase our samples beyond 80 words per sample. However, our results show that the TF-IDF performs better than the neural word embedding when used on short narratives (approximately 80 words per example in our case). It is in agreement with the comparison discussed in \cite{wang2017comparisons}. The difference in performance was less significant in our case. One can expect the neural word embeddings to outperform others approaches, when the word number increases as shown in \cite{sahlgren2016effects}.

Besides, with the same learning presentation approach (BoW, TF-IDF, or neural word embeddings), the LR classifiers had better performance than GaussianNB classifiers. The results align with the theoretical and experimental analysis from \cite{jordan2002discriminative, perlich2003tree}. LR performs better with smaller data sizes because it effectively approaches its lower asymptotic error from the initial learning steps. However, MLP-NN models always dominated with their best generalization. They have achieved their generalization capacity because the misclassification probability can be reduced and trained closer to optimal points that cannot be achieved with simple algorithms \cite{bartlett1998sample}. 

%, perlich2003tree

By applying the dropout (p=0.25) \cite{srivastava2014dropout}, GlorotNormal initializer \cite{glorot2010understanding}, and balancing the classes by using the Bayes Imbalance Impact Index \cite{lu2019bayes}, the classifier was successful in avoiding the overfitting. Primarily, Fig. \ref{fig:AUC} represents the Area Under the Curve (AUC) with respect to the epoch for the training and validation. We can see that the classifier can achieve nearly 100\% of separability of the two classes during the training. The classier can achieve almost 90\% of the separability during the validation. The distance between the two curves does not change with the increasing epoch number. And, the validation curve does not drop out to the growing epoch number. This indicates the algorithm does not overfit.

We also tested with the model CamemBERT, which is specifically a transformer-based language model for the french language \cite{martin2020camembert}. It is motivated by the success of a Bidirectional Encoder Representations from
Transformers (BERT) for natural language understanding \cite{devlin2019bert}. Unfortunately, the result was not as good as expected; we could only achieve less than 60\% accuracy, even though we applied the drop-out technique as recommended from the study \cite{pasupa2016comparison}. We continued investigating with the simpler Transformer, which is solely based on attention mechanisms through the connection of the encoder and decoder \cite{vaswani2017attention}, and it is implemented by Keras \cite{Transformer_ApoorvNandan}. The result has achieved a decent performance compared to advanced and complicated BERT-based models. However, it is still far below the performance from the simple MLP-NN, where the highest precision and recall are continually fluctuating at around 80\% as shown in Fig. \ref{fig:trans_recall}. Moreover, from the result of Fig. \ref{fig:trans_recall}, we can conclude that the transformer-based model underperforms in classification tasks for a small sample size, short of clinical NLP. This conclusion is in agreement with the limitations identified and discussed in \cite{gao2021limitations}; the authors have proved that the transformer-based model was well suited for understanding the contextual meaning of a long sequence rather than understanding key words or phrases.

\begin{table*}[!tp]
\footnotesize
\centering
\caption{Summarization of experiments performance evaluation}
\label{tab:table-results}
\resizebox{\textwidth}{!}{\begin{tabular}{|c|c|l|l|l|l|c|l|l|l|l|l|l|l|c|l|l|l|c|}
\hline
\multicolumn{2}{|c|}{\multirow{2}{*}{Representation}} &
  \multicolumn{1}{c|}{\multirow{2}{*}{ML}} &
  \multicolumn{4}{c|}{Exp\_1} &
  \multicolumn{4}{c|}{Exp\_2} &
  \multicolumn{4}{c|}{Exp\_3} &
  \multicolumn{4}{c|}{Exp\_4} \\ \cline{4-19} 
\multicolumn{2}{|c|}{} &
  \multicolumn{1}{c|}{} &
  acc &
  pre &
  rec &
  \multicolumn{1}{l|}{f1} &
  acc &
  pre &
  rec &
  f1 &
  acc &
  pre &
  rec &
  \multicolumn{1}{l|}{f1} &
  acc &
  pre &
  rec &
  \multicolumn{1}{l|}{f1} \\ \hline
\multirow{9}{*}{\rot{W/o Feature Selection}} &
  \multirow{3}{*}{BoW} &
  LR &
  0.80 &
  0.77 &
  0.81 &
  0.79 &
  0.81 &
  0.80 &
  0.82 &
  0.81 &
  0.82 &
  0.80 &
  0.83 &
  0.81 &
  0.82 &
  0.80 &
  0.81 &
  0.8 \\ \cline{3-19} 
 &
   &
  GaussianNB &
  0.77 &
  0.72 &
  0.80 &
  0.76 &
  0.78 &
  0.76 &
  0.81 &
  0.78 &
  0.79 &
  0.78 &
  0.81 &
  0.79 &
  0.79 &
  0.76 &
  0.80 &
  0.78 \\ \cline{3-19} 
 &
   &
  MLP-NN &
  0.81 &
  0.78 &
  0.81 &
  0.79 &
  0.81 &
  0.80 &
  0.82 &
  0.81 &
  0.81 &
  0.81 &
  0.84 &
  0.82 &
  0.81 &
  0.81 &
  0.82 &
  0.81 \\ \cline{2-19} 
 &
  \multirow{3}{*}{TF-IDF} &
  LR &
  0.81 &
  0.79 &
  0.82 &
  0.8 &
  0.78 &
  0.76 &
  0.79 &
  0.77 &
  0.81 &
  0.78 &
  0.81 &
  0.79 &
  0.77 &
  0.75 &
  0.77 &
  0.76 \\ \cline{3-19} 
 &
   &
  GaussianNB &
  0.79 &
  0.75 &
  0.81 &
  0.78 &
  0.77 &
  0.74 &
  0.80 &
  0.77 &
  0.78 &
  0.75 &
  0.81 &
  0.78 &
  0.76 &
  0.74 &
  0.80 &
  0.77 \\ \cline{3-19} 
 &
   &
  MLP-NN &
  0.81 &
  0.80 &
  0.82 &
  0.81 &
  0.84 &
  0.82 &
  0.85 &
  0.83 &
  0.85 &
  0.84 &
  0.85 &
  0.84 &
  0.82 &
  0.81 &
  0.81 &
  0.81 \\ \cline{2-19} 
 &
  \multirow{3}{*}{Embedding} &
  LR &
  0.74 &
  0.72 &
  0.79 &
  0.75 &
  0.76 &
  0.74 &
  0.79 &
  0.76 &
  0.78 &
  0.75 &
  0.82 &
  0.78 &
  0.76 &
  0.73 &
  0.77 &
  0.75 \\ \cline{3-19} 
 &
   &
  GaussianNB &
  0.72 &
  0.71 &
  0.77 &
  0.74 &
  0.76 &
  0.71 &
  0.79 &
  0.75 &
  0.76 &
  0.73 &
  0.80 &
  0.76 &
  0.75 &
  0.72 &
  0.72 &
  0.72 \\ \cline{3-19} 
 &
   &
  MLP-NN &
  0.74 &
  0.74 &
  0.76 &
  0.75 &
  0.77 &
  0.76 &
  0.78 &
  0.77 &
  0.79 &
  0.77 &
  0.80 &
  0.78 &
  0.77 &
  0.73 &
  0.78 &
  0.75 \\ \hline
\multirow{9}{*}{\rot{W/ Feature Selection}} &
  \multirow{3}{*}{BoW} &
  LR &
  0.80 &
  0.81 &
  0.78 &
  0.79 &
  0.81 &
  0.81 &
  0.79 &
  0.80 &
  0.78 &
  0.78 &
  0.77 &
  0.77 &
  0.80 &
  0.80 &
  0.79 &
  0.79 \\ \cline{3-19} 
 &
   &
  GaussianNB &
  0.80 &
  0.81 &
  0.78 &
  0.79 &
  0.80 &
  0.78 &
  0.79 &
  0.78 &
  0.78 &
  0.79 &
  0.77 &
  0.78 &
  0.80 &
  0.81 &
  0.78 &
  0.79 \\ \cline{3-19} 
 &
   &
  MLP-NN &
  0.80 &
  0.79 &
  0.80 &
  0.79 &
  0.82 &
  0.82 &
  0.81 &
  0.81 &
  0.83 &
  0.82 &
  0.83 &
  0.82 &
  0.84 &
  0.83 &
  0.84 &
  0.83 \\ \cline{2-19} 
 &
  \multirow{3}{*}{TF-IDF} &
  LR &
  0.76 &
  0.71 &
  0.79 &
  0.75 &
  0.82 &
  0.81 &
  0.83 &
  0.82 &
  0.83 &
  0.82 &
  0.83 &
  0.82 &
  0.78 &
  0.78 &
  0.80 &
  0.79 \\ \cline{3-19} 
 &
   &
  GaussianNB &
  0.80 &
  0.78 &
  0.80 &
  0.79 &
  0.81 &
  0.82 &
  0.79 &
  0.80 &
  0.81 &
  0.81 &
  0.82 &
  0.81 &
  0.79 &
  0.78 &
  0.79 &
  0.78 \\ \cline{3-19} 
 &
   &
  MLP-NN &
  \textbf{0.84} &
  \textbf{0.84} &
  \textbf{0.85} &
  \textbf{0.84} &
  \textbf{0.87} &
  \textbf{0.86} &
  \textbf{0.88} &
  \textbf{0.87} &
  \textit{\textbf{0.89}} &
  \textit{\textbf{0.89}} &
  \textit{\textbf{0.88}} &
  \textit{\textbf{0.88}} &
  \textbf{0.85} &
  \textbf{0.84} &
  \textbf{0.84} &
  \textbf{0.84} \\ \cline{2-19} 
 &
  \multirow{3}{*}{Embedding} &
  LR &
  0.80 &
  0.78 &
  \multicolumn{1}{c|}{0.80} &
  0.79 &
  \multicolumn{1}{c|}{0.80} &
  \multicolumn{1}{c|}{0.79} &
  \multicolumn{1}{c|}{0.80} &
  \multicolumn{1}{c|}{0.79} &
  \multicolumn{1}{c|}{0.82} &
  \multicolumn{1}{c|}{0.82} &
  \multicolumn{1}{c|}{0.83} &
  0.82 &
  \multicolumn{1}{c|}{0.81} &
  \multicolumn{1}{c|}{0.78} &
  \multicolumn{1}{c|}{0.79} &
  0.78 \\ \cline{3-19} 
 &
   &
  GaussianNB &
  0.77 &
  0.76 &
  \multicolumn{1}{c|}{0.78} &
  0.77 &
  \multicolumn{1}{c|}{0.79} &
  \multicolumn{1}{c|}{0.79} &
  \multicolumn{1}{c|}{0.79} &
  \multicolumn{1}{c|}{0.79} &
  \multicolumn{1}{c|}{0.81} &
  \multicolumn{1}{c|}{0.81} &
  \multicolumn{1}{c|}{0.80} &
  0.8 &
  \multicolumn{1}{c|}{0.79} &
  \multicolumn{1}{c|}{0.78} &
  \multicolumn{1}{c|}{0.78} &
  0.78 \\ \cline{3-19} 
 &
   &
  MLP-NN &
  0.80 &
  0.79 &
  \multicolumn{1}{c|}{0.80} &
  0.79 &
  \multicolumn{1}{c|}{0.80} &
  \multicolumn{1}{c|}{0.80} &
  \multicolumn{1}{c|}{0.80} &
  \multicolumn{1}{c|}{0.80} &
  \multicolumn{1}{c|}{0.82} &
  \multicolumn{1}{c|}{0.81} &
  \multicolumn{1}{c|}{0.81} &
  0.81 &
  \multicolumn{1}{c|}{0.80} &
  \multicolumn{1}{c|}{0.79} &
  \multicolumn{1}{c|}{0.80} &
  0.79 \\ \hline
\end{tabular}}
\vspace{-3mm}
\end{table*}

\section{Conclusion}
\label{sec:conclusion}

We have employed both learning representation and machine learning algorithms to tackle the French clinical natural language processing for detecting cardiac failure in children at CHUSJ. We have extensively conducted and analyzed a conceptual framework to detect a patient's health condition from the contextual input to the contextual output. Our numerical results have confirmed the feasibility of the proposed design by combining TF-IDF and MLP-NN; the proposed mechanism could also be improved with the feature selection from the learning representation vector space. Consequently, the proposed framework yields an overall classification performance with 89\% accuracy, 88\% recall,  and 89\% precision.

Secondly, we assumed that the numeric values significantly contribute to the classifier. Instead of losing them, we addressed different decoding approaches for numeric values in our work. In our case study, encoding the decimal point into a string “DOT” has helped the learning representation process retain the information from the numerical values in clinical notes. Otherwise, it is better to remove the numeric values rather than keep them without any encoding, or extensive encoding. 

Finally, with the MLP-NN learning algorithm, we can train closer to optimal architectures, which cannot be trained with simple algorithms (LR, GaussianNB, RF, MultinomialNB, and SVM). Although BERT-based models are currently known as the state-of-the-art in natural language processing tasks, the final results suggest that these Transformer-based methods perform less effectively than existing alternatives. 

One of the limitations is that the CDSS is still under development (in process currently). The next step of our project is to create the CDSS to diagnose ARDS early by integrating this NLP algorithm with the other algorithms on hypoxemia and chest X-Ray analysis. When the integration is done in the PICU electronic medical infrastructure, we will validate the CDSS's ability to screen ARDS prospectively. Furthermore, future research should carefully consider the potential effects of numerical values alongside unstructured notes. Ideally, an algorithm, which can automatically extract and represent the numerical values from the clinical notes, should be investigated for further validation. This may be a promising aspect of using a semantic neural network to determine the boundaries and extract the numerical values from the text. And, generative learning has a great potential for an evaluation \cite{dua2019drop}. 

\section*{Acknowledgment}

The clinical data were provided by the Research Center at CHU Sainte-Justine hospital. The authors thank Dr. Sally Al Omar, Dr. Rambaud Jérôme and Dr. Sans Guillaume for their data support of this research. This work was supported by a scholarship from Fonds de recherche du Québec – Nature et technologies (FRQNT) to Thanh-Dung Le, the funds from Natural Sciences and Engineering Research Council (NSERC) and Institut de valorisation des données (IVADO), and the Fonds de la recherche en sante du Quebec (FRQS).

\section*{Supplementary Materials}
The additional exploratory data analysis from the data pre-processing step is presented in supplementary materials. We also provide an overview theoretical explanation of employed methods for clinical note representation learning, machine learning classifiers, and imbalance learning in that document.

%\vspace{140pt}
%\balance
\bibliographystyle{IEEEtran}
\bibliography{IEEEabrv,Bibliography}

\newpage


\section*{Supplementary material}

# Supplementary Materials

Detecting of a Patient's Condition From Clinical Narratives Using Natural Language Representation

Thanh-Dung Le, *Member, IEEE,* Rita Noumeir Ph.D., *Member, IEEE,* Jérôme Rambaud M.D., Ph.D., Guillaume Sans M.D., and Philippe Jouvet M.D., Ph.D.

**T**HIS Supplementary Materials provides additional analysis for the exploratory data analysis of the data pre-processing step. It also provides an overview theoretical explanation of employed methods for clinical note representation learning, machine learning classifiers, and imbalance learning.

## I. Exploratory Data Analysis

Stop-word removal is an essential step to exclude the minor information. Fig. 1 shows the statistics for the list of stop words removed. These are all the definition in French, including de and d' (of, to, from, by, with, at, off), et (and), à (to, in, at, with, by, upon), avec (with), pas (not, without), en (in, into, to, of, at, during), la and l'(the French definite articles).

For the note visualization, we apply the ScatterText [1]. We have more than 580000 unigrams (n-gram) shown in Fig. 2. The figure shows the most frequent words for the positive case in the upper right corner; the most frequent words for the negative cases in the lower-left corner; and, all less frequent words for both cases are in the center. Besides, the top terms from the positive and negative cases are presented on the right-hand side. In positive cases, we quickly see that most of these terms are positively related to cardiac malfunction: milrinone or milri (milrinone), aorte or aortique valve (aortic valve). In contrast, terms such as respiratoire (respiratory), and IVRS (Infection des voies respiratoires supérieures - Virus responsible for respiratory distress) indicate respiratory syndromes.

All the notes are short narratives, and the n-gram length distribution is shown in Fig. 3. The longest n-gram is over 400 words, but most of the n-gram length distribution is between 50 and 125 words. Besides, the top 20 terms from the positive and negative cases and their frequency distribution are illustrated

| term | Positive freq | Negative freq | pos_precision | pos_freq_pct | pos_hmean |
|---|---|---|---|---|---|
| de | 11460 | 13500 | 0.459135 | 0.043645 | 0.079712 |
| et | 5180 | 5519 | 0.484157 | 0.019728 | 0.037911 |
| à | 4437 | 6803 | 0.394751 | 0.016898 | 0.032409 |
| avec | 3818 | 4680 | 0.449282 | 0.014541 | 0.028169 |
| pas | 3449 | 5335 | 0.392646 | 0.013135 | 0.025420 |
| d | 3381 | 4191 | 0.446513 | 0.012876 | 0.025031 |
| en | 2957 | 3250 | 0.476398 | 0.011262 | 0.022003 |
| la | 2577 | 3404 | 0.430864 | 0.009814 | 0.019191 |
| l | 2425 | 3655 | 0.398849 | 0.009235 | 0.018053 |
| 0 | 2343 | 1055 | 0.689523 | 0.008923 | 0.017618 |

Fig. 1. Clinical notes analyzing for stop words removing.

in Fig. 4 and Fig. 5. There is an abbreviation for medical terms, whose descriptions and characteristics are summarized in Table I. For example, there are CIV (Communication intraventriculaire), CEC (Circulation extra-corporelle), CIA (Communication intra-auriculaire), FC (fréquence cardiaque), and IVRS (Infection des voies respiratoires supérieures).

For more specific, Fig. 6 shows the top 30 n-grams that frequently appear for both cases, and Fig. 7 also visualizes the terms distributed for both cases. In positive cases, we quickly see that most of these terms are positively related to cardiac malfunction: milrinone (milri), aorte, aortique, and valve. In contrast, terms such as *urgence, ad, respiratoire, vers, toux, and ivrs* indicate respiratory syndromes. From Fig. 6, we can see that the overlapping: terms that strongly indicate one class also appear to a lesser degree in the other class.

## II. Clinical Natural Language Representation Learning

We will evaluate the effectiveness of two conventional representation approaches, including Bag-of-Words, term frequency (TF) × inverse document frequency (IDF), and the word2vec neural embedding model.

### A. Bag-of-Words

A bag-of-words model [2], or BoW for short, is a representation of text that describes the occurrence of words within a document. Technically, if documents have similar content, they will be similar. It extracts features from the text by considering each word count as a feature from a vocabulary of known words. There is a theoretical analysis for understanding the BoW model [3], which proves that the success of BoW representation is not by using a heuristic clustering process but by a statistical approach and based on statistical consistency. However, in the BoW representation, any two different words drawn from the vocabulary are treated equally if they are assigned with the same topic; in reality, it neglects much correlation information among words.

### B. TF-IDF

The TF-IDF, first introduced in [4], stands for term frequency (TF) × inverse document frequency (IDF). TF-IDF weighting is commonly used in information retrieval for text mining. The intuition is that term importance increases with the term's frequency in the text, but its frequency neutralizes it in the domain of interest. Given a collection of terms $t \in T$ that appear in a set of $N$ documents $d \in D$, each of length $n_d$, tf-idf weighting is computed as follows:

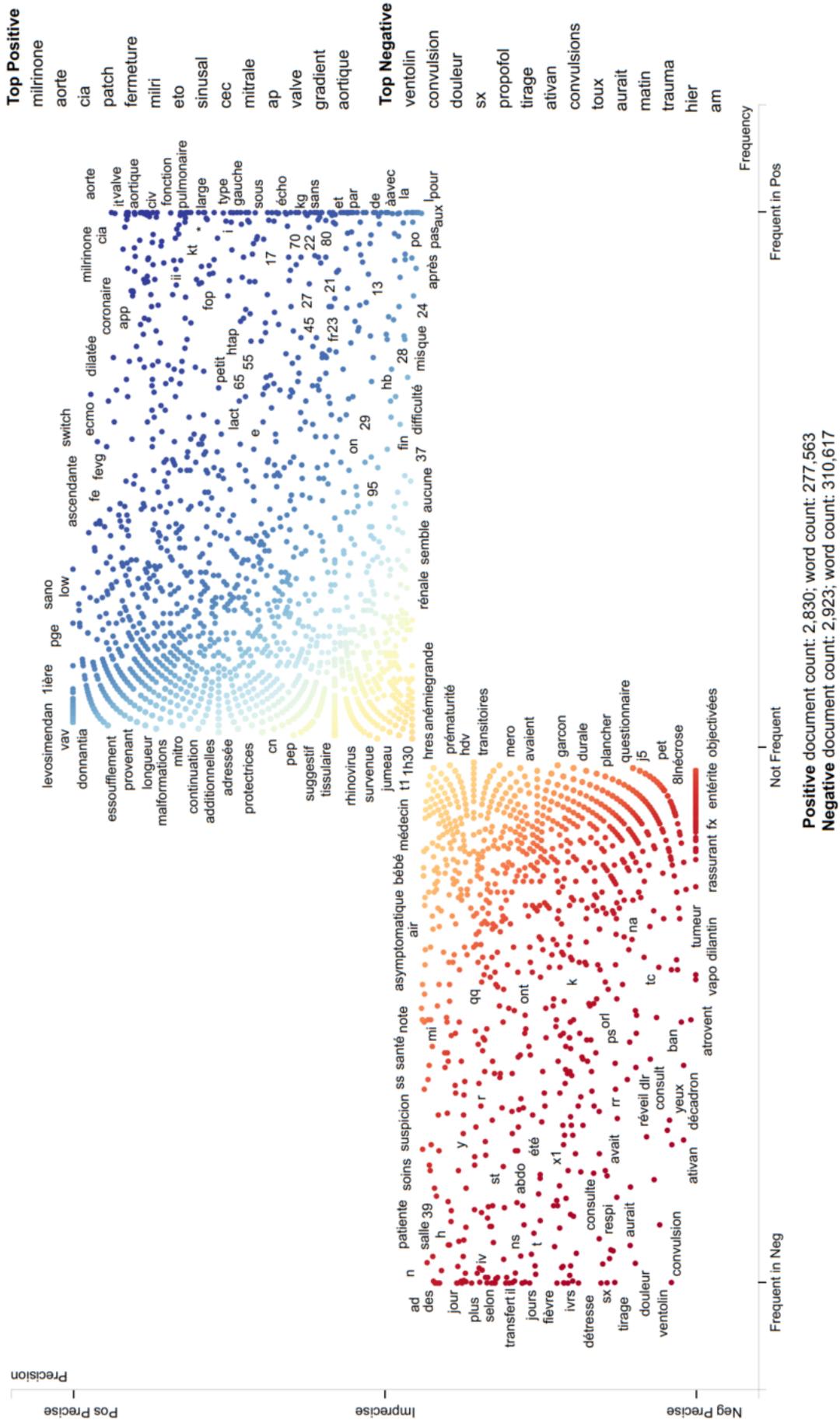

Fig. 2. Clinical note illustration by using Scattertext visualization.

TABLE I. Important Abbreviations for Medical Terms

| Abbreviations | Descriptions (In French) | Characteristics |
|---|---|---|
| CIV | Communication intraventriculaire | Cardiac malformation |
| CEC | Circulation extracorporelle | Treatment for cardiac failure |
| CIA | Communication intraauriculaire | Cardiac malformation |
| FC | Fréquence cardiaque | Cardiac frequency |
| IVRS | Infection des voies respiratoires supérieures | Virus responsible for respiratory distress |
| SOP | Salle d'opération | Operations |
| PO | Per os (by mouth) | Feeding |

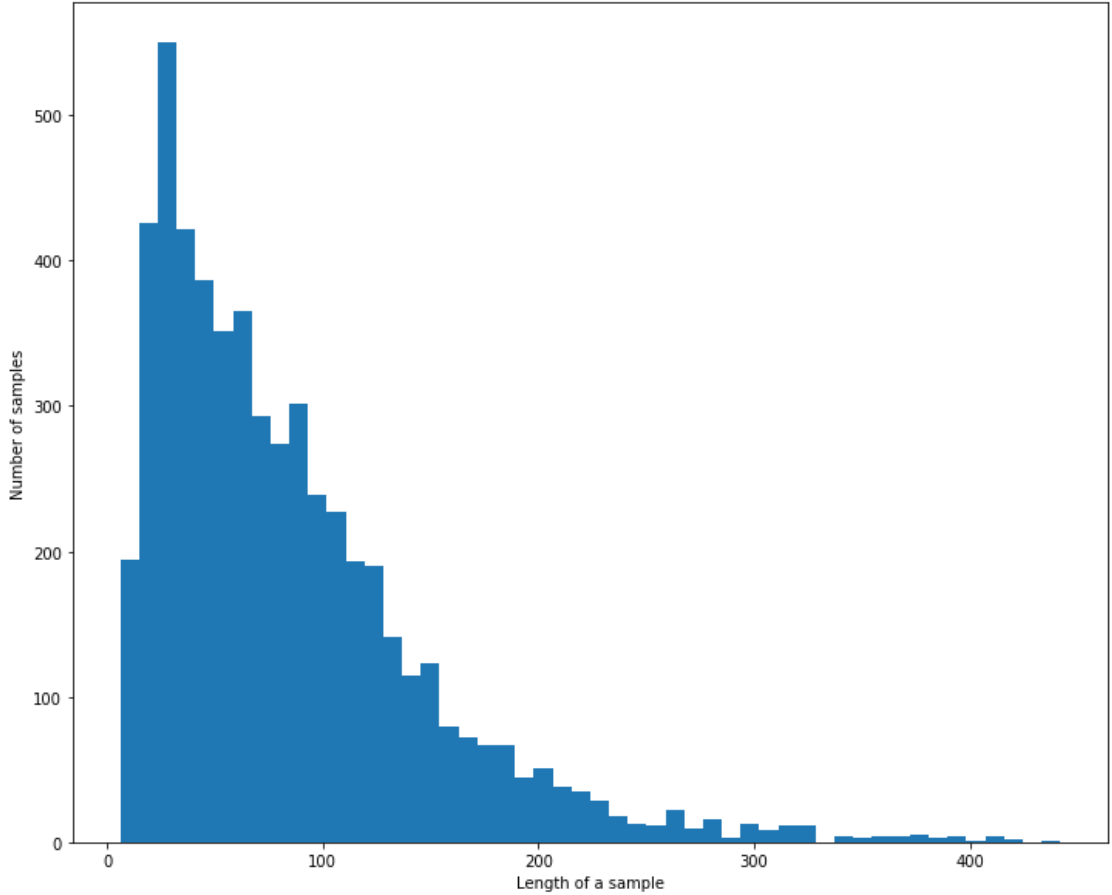

Fig. 3. The distribution of length of notes in the CHUSJ dataset.

$$tf_{t,d} = \frac{f_{t,d}}{n_d} \quad (1)$$

$$idf_t = \log \frac{N}{df_t} \quad (2)$$

$$W_{t,d} = tf_{t,d} \times idf_t \quad (3)$$

where $f_{t,d}$ is the frequency of term $t$ in document $d$, and $df_t$ is the document frequency of term $t$, the number of documents in which term $t$ appears. Several variations were offered, including normalizing $tf_{t,d}$, and optional weighting schemes.

TF-IDF and its variations do not only focus on the fundamental statistical relation of note representation. But, its strong theoretical arguments are explained for its heuristic [5], and as a probabilistic theoretical explanation [6]. And, it is shown that if TF-IDF is treated carefully with the bias, it will manifest better performance in representation learning.

### C. Neural Word Embeddings

The neural word embedding was introduced in [7], and named the word2vec model. Each word will be represented by two sets of vectors, $u_w$, and $v_w$. $u_w$ is used when word $w$ is the context word, and $v_w$ is used when word $w$ is the center word. Using these two vectors, the probability for the central target word $w_o$ and context word $w_c$ for each word $w$ will look like this:

$$p(w_o \mid w_c) = \frac{\exp(u_o^\top v_c)}{\sum_{i \in V} \exp(u_i^\top v_c)} \quad (4)$$

The key to the success of word2vec is that it can compute the logarithmic conditional probability for the central word vector and the context word vector. Then, its computation obtains the conditional likelihood for all the words in the dictionary given the context word $w_c$, and is trained by a neural network.

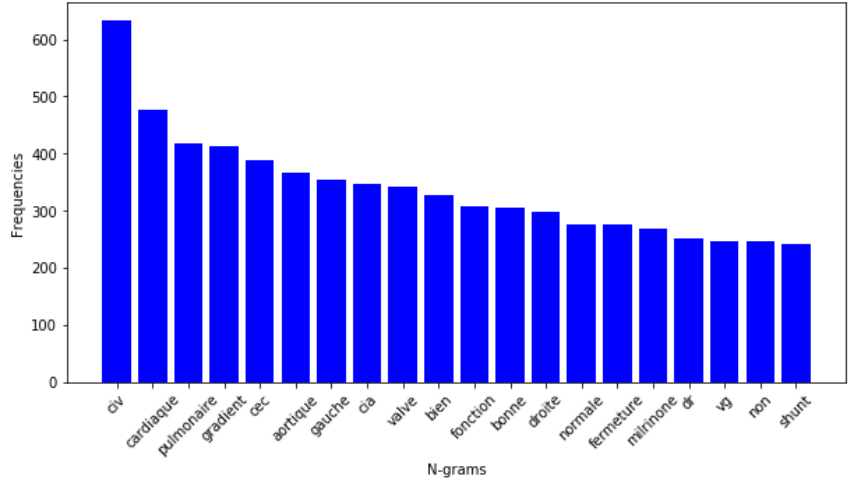

Fig. 4. N-grams's frequency distribution for positive cases (Top 20 n-grams).

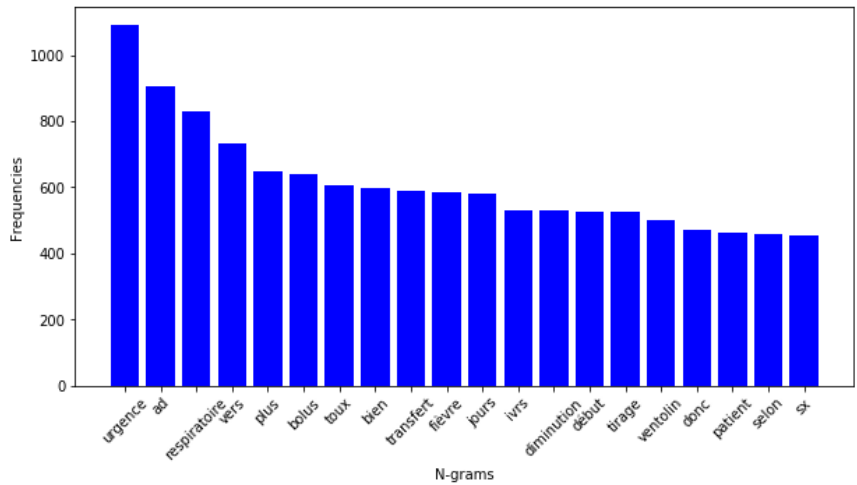

Fig. 5. N-grams's frequency distribution for negative cases (Top 20 n-grams).

$$\log p(w_o \mid w_c) = u_o^\top v_c - \log \left( \sum_{i \in V} \exp(u_i^\top v_c) \right) \quad (5)$$

where $V = \{0, 1, \ldots, |V|-1\}$ is the vocabulary index. After the training, for any word in the dictionary with index $w_i$, we will get its two-word vector sets $v_i$ and $u_i$. In applications of NLP, the central target word vector in the skip-gram model is generally used as the representation vector of a word.

## III. MACHINE LEARNING CLASSIFIERS

Here we consider a binary classification problem. Given $n$ training samples $D = (x_1, y_1)...,(x_n, y_n)$ where $x_i \in \mathcal{R}^p$ is a p-dimensional column vector and label $y_i \in \{0, 1\}$.

### A. Logistic Regression (LR)

LR uses a logistic function to model the probability of a binary dependent variable (particular class or event) such as unhealthy/healthy. Therefore, it is widely used in most medical fields [8]. We first write the logistic function as follows:

$$p(x; w) = p(y = 1 \mid x; w) = \frac{1}{1 + \exp(-w^T x)} \quad (6)$$

where $w$ is the weight vector of coefficients, and $p(\cdot)$ is a sigmoid function. Here, we assume that the $n$ training examples are generated independently. Finally, we thus can obtain the following log-likelihood.

$$\begin{aligned} \ell(w) &= \sum_{i=1}^{n} \log p(y_i \mid x_i; w) \\ &= \sum_{i=1}^{n} \left\{ y_i w^T x_i - \log \left( 1 + \exp \left( w^T x_i \right) \right) \right\} \end{aligned} \quad (7)$$

### B. Gaussian Naïve Bayes (GaussianNB)

GaussianNB algorithm for classification is a set of supervised learning algorithms based on applying Bayes' theorem with the "Naïve" assumption of conditional independence between every pair of features given the value of the class variable. Bayes' theorem states the following relationship,

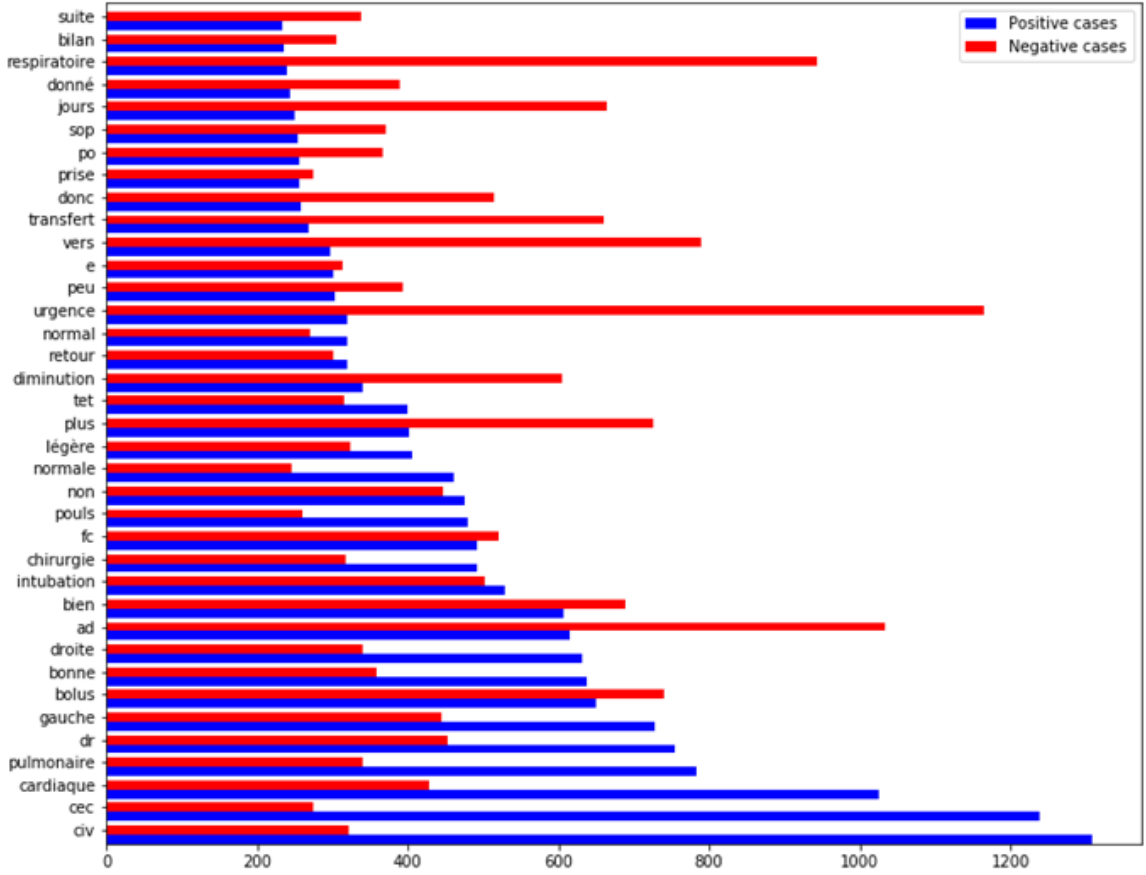

Fig. 6. Top 30 frequent n-grams overlapping respecting to both two classes distribution.

given class variable $Y$ and dependent feature vector $x_1$ through $x_n$:

$$p(y \mid x_1, \ldots, x_n) = \frac{p(y) p(x_1, \ldots, x_n \mid y)}{p(x_1, \ldots, x_n)} \quad (8)$$

Using the Naïve conditional independence assumption that

$$p(x_i \mid y, x_1, \ldots, x_{i-1}, x_{i+1}, \ldots, x_n) = p(x_i \mid y), \quad (9)$$

for all $i$, this relationship is simplified to

$$p(y \mid x_1, \ldots, x_n) = \frac{p(y) \prod_{i=1}^{n} p(x_i \mid y)}{p(x_1, \ldots, x_n)} \quad (10)$$

Since $p(x_1, \ldots, x_n)$ is constant given the input, we can use the following classification rule: $P(y \mid x_1, \ldots, x_n) \propto p(y) \prod_{i=1}^{n} p(x_i \mid y)$. Then

$$\hat{y} = \arg\max_{y} p(y) \prod_{i=1}^{n} P(x_i \mid y) \quad (11)$$

and the maximum a posteriori estimation can be used to estimate $p(y)$ and $p(x_i \mid y)$; the former is the relative frequency of class $y$ in the training set. The difference between GaussianNB and Naïve Bayes classifiers is mainly based on $p(x_i \mid y)$ distribution. The likelihood of the features is assumed to be Gaussian with mean $\mu$ and variance $\sigma^2$ for GaussianNB [9]:

$$p(x_i \mid y) = \frac{1}{\sqrt{2\pi\sigma_y^2}} \exp\left(-\frac{(x_i - \mu_y)^2}{2\sigma_y^2}\right) \quad (12)$$

### C. Multilayer Perceptron Neural Network (MLP-NN)

The earliest work in the field of "neural network" is attempted to understand, model, and emulate neurological function and learning in brains [10]. Since then, a commonly used neural network has been the MLP-NN. In an MLP-NN, neurons are structured into layers, consisting of at least three layers: the input layer, the hidden layer or layers, and an output layer [11]. Typical MLP-NN networks are feedforward neural networks where the computation is carried out in a single direction from input to output.

Typically, MLP-NN are feedforward neural networks where $x_i$ is the $i^{th}$ input (axon from a neuron) to an input neuron, and a weight $w_i$ is the effect of the $i^{th}$ synapse on the neural network. Then, the total impact of the input on the synapse of the cell body is:

$$\sum_{i}^{n} w_i x_i + b \quad (13)$$

Following this affine transformation of the weighted sum of its input, a non-linear activation function $f(\cdot)$ is defined

Fig. 7. Visualization of terms distribution for both classes.

for the cell output. This non-linear activation function $f(\cdot)$ generally enables the MLP-NN to learn and solve non-linear problems. It is proven that ReLU activation in combination with stochastic gradient descent optimization algorithm shows a better convergence and gives a better optimization in small MLP-NN [12]. Therefore, we apply the ReLU activation for the hidden layer as shown below:

$$f(\cdot) = ReLU(\alpha) = \max(\alpha, 0) \qquad (14)$$

## IV. Imbalance Learning

The number of positive and negative cases are not symmetric; hence, to deal with that imbalance problem, we apply the balance technique Bayes Imbalance Impact Index ($BI^3$) for the classifiers [13]. An evaluation of different sampling techniques and neural network architectures recently confirmed that the effect of sampling and the neural network architecture strongly depend on the data set. The results become less consistent [14]. Consequently, we considered two reasons for not being interested in sampling approaches (i) characteristics of our clinical notes data: small training size, a large number of features, and correlations between features. Also, it is concluded that there is no clear winner between oversampling and undersampling to compensate for the class imbalance if we do not carefully take into account factors such as class distribution, class prevalence, and features correlations in medical decision making [15]; (ii) sampling methods focused only on the imbalance ratio but ignored the lack of information for the minority class, overlap, and disjuncts, in addition to the lack of representation, which cannot be learned because it is not present in the training set [16]. In summary, the $BI^3$ for the whole data set $\mathcal{D}$ is calculated by averaging over all Individual Bayes Imbalance Impact Index $IBI^3$ on the minority class.

$$BI^3(\mathcal{D}) = \frac{1}{N_p} \sum_{(x_i, y_i) \in \mathcal{D}, y_i = +1} IBI^3(x_i) \qquad (15)$$

$N_p$ is the number of samples in the positive class, and $IBI^3$ is the difference between the normalized posterior probabilities of the imbalanced and balanced case.

$$IBI^3 = \frac{f'_p(x)}{f_n(x) + f'_p(x)} - \frac{f_p(x)}{f_n(x) + f_p(x)} \qquad (16)$$

$f_p(x)$ and $f_n(x)$ are the posterior scores for the positive and negative classes, respectively. $f'_p(x)$ is proportional to the posterior probability of the minority class in the estimated balanced case. If $IBI^3$ is small, the influence of the imbalance on $x$ is irrelevant. Therefore, it can measure how the imbalance influences a single minority class sample.



\end{document}